\documentclass[review]{elsarticle}

\usepackage[normalem]{ulem}
\usepackage[table,xcdraw]{xcolor}
\usepackage{soul}
\usepackage{colortbl}
\usepackage{makecell}
\usepackage{url}
\usepackage{multirow}
\usepackage[section]{placeins}
\usepackage{svg}
\usepackage{graphicx}
\usepackage{caption}
\usepackage{subcaption}
\usepackage{float}
\usepackage{bm}
\usepackage{mathtools}
\usepackage{breqn}
\usepackage{fontawesome}
\usepackage{pdflscape}

\usepackage[numbers]{natbib}

\usepackage{tikz}
\usetikzlibrary{spy,backgrounds,calc,shapes}

\usepackage{etoolbox}
\AtBeginEnvironment{algorithm}{\linespread{1.4}\selectfont}

\newenvironment{conditions*}
  {\par\vspace{\abovedisplayskip}\noindent
   \tabularx{\columnwidth}{>{$}l<{$} @{${}-{}$} >{\raggedright\arraybackslash}X}}
  {\endtabularx\par\vspace{\belowdisplayskip}}

\makeatletter
\AtBeginDocument{%
  \expandafter\renewcommand\expandafter\subsection\expandafter{%
    \expandafter\@fb@secFB\subsection
  }%
}
\makeatother

\begin{document}

\begin{frontmatter}

\title{Squeezing nnU-Nets with Knowledge Distillation for On-Board Cloud Detection}

\author{Bartosz Grabowski$^{1}$}
\author{Maciej Ziaja$^{1,2}$}
\author{Michal Kawulok$^{1,2}$}
\author{Piotr Bosowski$^{1}$}
\author{Nicolas Long\'{e}p\'{e}$^{3}$}
\author{Bertrand Le Saux$^{3}$}
\author{Jakub Nalepa$^{1,2}$}\ead{Jakub.Nalepa@polsl.pl}\corref{mycorrespondingauthor}
\address{
$^1$KP Labs, Konarskiego 18C, 44-100 Gliwice, Poland\\
$^2$Department of Algorithmics and Software, Silesian University of Technology, Akademicka 16, 44-100 Gliwice, Poland\\
$^3$European Space Agency, Largo Galileo Galilei 1, 00044 Frascati, Italy\\
}
\cortext[mycorrespondingauthor]{Corresponding author}

\begin{abstract}

Cloud detection is a pivotal satellite image pre-processing step that can be performed both on the ground and on board a satellite to tag useful images. In the latter case, it can reduce the amount of data to downlink by pruning the cloudy areas, or to make a satellite more autonomous through data-driven acquisition re-scheduling. We approach this task with nnU-Nets, a self-reconfigurable framework able to perform meta-learning of a segmentation network over various datasets. Unfortunately, such models are commonly memory-inefficient due to their (very) large architectures. To benefit from them in on-board processing, we compress nnU-Nets with knowledge distillation into much smaller and compact U-Nets. Our experiments, performed over Sentinel-2 and Landsat-8 images revealed that nnU-Nets deliver state-of-the-art performance without any manual design. Our approach was ranked within the top 7\% best solutions (across 847 teams) in the On Cloud N: Cloud Cover Detection Challenge, where we reached the Jaccard index of 0.882 over more than 10k unseen Sentinel-2 images (the winners obtained 0.897, the baseline U-Net with the ResNet-34 backbone: 0.817, and the classic Sentinel-2 image thresholding: 0.652). Finally, we showed that knowledge distillation enables to elaborate dramatically smaller (almost 280$\times$) U-Nets when compared to nnU-Nets while still maintaining their segmentation capabilities.

\end{abstract}

\end{frontmatter}


\section{Introduction} \label{sec:intro}

Detecting clouds in satellite images plays a key role in the pre-processing chain of such imagery~\cite{JEPPESEN2019247}, especially given that the global cloud coverage is about 68\% annually~\cite{Li2021}. Understanding which parts of the scene are cloudy can allow to not only reduce the amount of data for further on-the-ground processing, but may also help re-schedule the  acquisition process of the particularly important areas to  retrieve clean images. Additionally, the cloud coverage may bring additional information concerning climate change, hurricanes, volcanic activity, and many more events that can be observed from space~\cite{38-cloud-1}. Therefore, developing accurate cloud detection algorithms is of paramount importance nowadays to optimize the mission operations through appropriately handling the low-quality or useless (i.e.,~fully covered by clouds) image data~\cite{Mahajan2020}.

Capturing large, representative, and heterogeneous annotated cloud detection sets is cumbersome and difficult in practice, as they should reflect various factors that affect the satellite image characteristics, such as atmospheric distortions, latitude, ground reflectance, and other~\cite{rs13081532}. Despite those challenges, new sets have been emerging---they are commonly created in a semi-automated way with certain quality procedures adopted to ensure sufficient ground-truth (GT) quality. Such benchmarks can be used for training and verifying cloud detection, and they include the cloud masks for Sentinel-2 (S-2) images~\cite{rs13204100}, the S-2 set released in the On Cloud N Challenge\footnote{This set is available at \url{https://mlhub.earth/data/ref_cloud_cover_detection_challenge_v1} (accessed on June 10, 2023).}, or the Landsat-8 (L-8) 38-Cloud set~\cite{38-cloud-1,38-cloud-2}.

Classic cloud detection techniques based on rule-based approaches and time differentiation methods~\cite{Mahajan2020} are time-efficient and trivial to implement. However, they generalize poorly across different sensors and acquisition conditions, suffer from thin cloud omission and non-cloud bright pixel commission, and are heavily based on the prior knowledge about the cloud characteristics. Mohajerani and Saeedi designed a specialized Cloud-Net deep learning architecture for L-8 images~\cite{38-cloud-1} which outperformed a more generic U-Net~\cite{38-cloud-2} and the improved Fmask algorithm fine-tuned for L-8~\cite{ZHU2015269} (originally developed for Landsats 4--7). Those techniques required manual redesign to process L-8, and would likely need a similar procedure for other satellites, which reduces their flexibility. Similarly, Domnich et al.~utilized U-Nets for detecting clouds in S-2 images~\cite{rs13204100}, and Yanan et al.~enhanced U-Nets with attention modules for L-8~\cite{Yanan_2020}. All of the above-mentioned algorithms were manually designed by humans---it is commonly a trial-and-error and time-consuming procedure. To tackle it, various ML techniques for automated network architecture optimization (AutoML) were introduced for different tasks, also including satellite image classification~\cite{10.1007/978-3-030-86517-7_28}. Such approaches have not been utilized for the fundamental task of cloud detection. We address this research gap to make the deployment of the cloud detection algorithms as smooth as possible for emerging missions.

Developing resource-frugal machine learning models is of the highest urgency to deploy them on edge devices, e.g.,~an imaging satellite, to bring ``the brain close to the eyes''. This approach allows for processing raw data on board a satellite to download actionable items rather than raw data itself for further analysis, and to provide inherent scalability of Earth observation solutions. If a model is to be uplinked to a reconfigurable satellite, reducing its size will directly affect the bandwidth requirements~\cite{rs13193981}. We tackle the issue of developing memory-efficient yet well-performing models, and exploit knowledge distillation (KD) for this task~\cite{Gou2021}. We hypothesize that exploiting not only the GT data but also the features of AutoML models during the training process can enhance the capabilities of much smaller deep models.

Our contribution is two-fold: (\textit{i})~we introduce a fully data-driven AutoML-powered approach for cloud detection in satellite image data which requires zero user intervention, and (\textit{ii})~propose to benefit from such models while developing compact U-Nets through exploiting KD. We build upon the nnU-Net (which unfolds to no-new-U-Net) framework that has already established the state of the art in an array of medical image analysis tasks, including the segmentation of brain tumors, liver, prostate, spleen or kidney~\cite{isensee_nnu-net_2021}. In the hands-free processing chain, the nnU-Net adapts its most important components according to the input image data, hence the resulting deep learning model, together with the pre- and post-processing routines, data augmentation and training settings are influenced by the characteristics of the training set. This opens new doors for deploying such cloud detection engines in emerging applications based solely on the available image data, without the need to manually redesign the algorithms. Our experiments performed over two datasets captured by different missions (Landsat-8 and Sentinel-2) showed that the nnU-Nets deliver segmentation competitive with the state of the art (we were ranked in the top 7\% teams within the On Cloud N: Cloud Cover Detection Challenge with almost 850 participating teams). Since nnU-Nets are often extremely memory-hungry, we propose to utilize KD to train much smaller U-Nets while still benefiting from the characteristics of well-performing AutoML models, and validate them over the simulated $\Phi$-Sat-2 image data. Finally, the elaborated architectures, together with their parameterization, and our code for preparing the L-8 and S-2 images for nnU-Nets are available at \url{https://gitlab.com/jnalepa/nnUNets_for_clouds}.

\section{Materials and Methods}\label{sec:method}

We discuss the data used in our study in Sect.\,\ref{sec:datasets}. To show flexibility, we exploit both S-2 and L-8 images. The nnU-Nets for segmenting clouds are presented in Sect.\,\ref{sec:method}.

\subsection{Description of Datasets}\label{sec:datasets}

The \textbf{38-Cloud (38-C)} dataset contains 38 L-8 images and their pixel-level GT cloud segmentations. The images are cropped into $384\times384$ patches by the dataset's authors~\cite{38-cloud-1,38-cloud-2}, with 8,400 training and 9,201 test patches. In 38-Cloud, thin clouds (haze) are also annotated as clouds. The \textbf{On Cloud N: Cloud Cover Detection Challenge (OCN)} dataset consists of S-2 imagery divided into 11,748 training patches, collected between 2018 and 2020, and 10,980 test patches (all of them are of $512\times 512$ pixels, 10\,m spatial resolution). Each patch was captured for a specific area, mostly in Africa, South America, and Australia. Although the organizers included four bands (B02, B03, B04, and B08), the available S-2 database could be used to pull all missing bands. We exploited the bands suggested by the organizers only to follow their anticipated scenario. The labels for OCN were manually generated using the optical S-2 bands. Finally, we use the \textbf{KappaZeta (KZ)} dataset, being one of the most comprehensive nowadays~\cite{rs13204100} in the KD experiment. It consists of 4,403 scenes (512$\times$512) of the Northern European terrestrial area. KZ is a six-class set: clear pixel (not obscured by clouds, 35.1\% of all pixels), thick clouds (31.7\%), semi-transparent clouds (17.7\%), cloud shadow (7.7\%), undefined (pixels with uncertain labels, 6.5\%), and missing (corrupted pixels, 1.4\%). Here, we emulate the expected $\Phi$-Sat-2 data based on the original S2 imagery: top-of-atmosphere reflectance was changed into spectral radiance, and we simulated the target Simera sensor characteristics, including the point spread function and the signal-to-noise ratio for each band independently~\cite{rs13081532}. See other details of the datasets in Table~\ref{tab:datasets}.

\begin{table}[ht!]
\scriptsize
\centering
  \caption{The characteristics of the 38-Cloud (L-8) and OCN (S-2) datasets.}
  \label{tab:datasets}
  \renewcommand{\tabcolsep}{1.8mm}
  \begin{tabular}{ccccccccccc}
    \Xhline{2\arrayrulewidth}
    \multicolumn{2}{c}{Band ID} && \multicolumn{2}{c}{Band name} && \multicolumn{2}{c}{Wavelength [nm]} && \multicolumn{2}{c}{Resolution [m]}\\
    \cline{1-2} \cline{4-5} \cline{7-8} \cline{10-11}
    38-C & OCN && 38-C & OCN && 38-C & OCN && 38-C & OCN \\
    \hline
    B02 & B02 && \multicolumn{2}{c}{Blue} && 450--515 & 458--523 && 30 & 10\\
    B03 & B03 && \multicolumn{2}{c}{Green} && 520--600 & 543--578 && 30 & 10\\
    B04 & B04 &&  \multicolumn{2}{c}{Red} && 630--680 & 650--680 && 30 & 10\\
    B05 & B08 &&  \multicolumn{2}{c}{NIR} && 845--885 & 785--899 && 30 & 10\\
  \Xhline{2\arrayrulewidth}
\end{tabular}
\end{table}

\subsection{Proposed Method}\label{sec:method}

\begin{figure}[ht!]
     \centering
     \includegraphics[width=0.78\columnwidth]{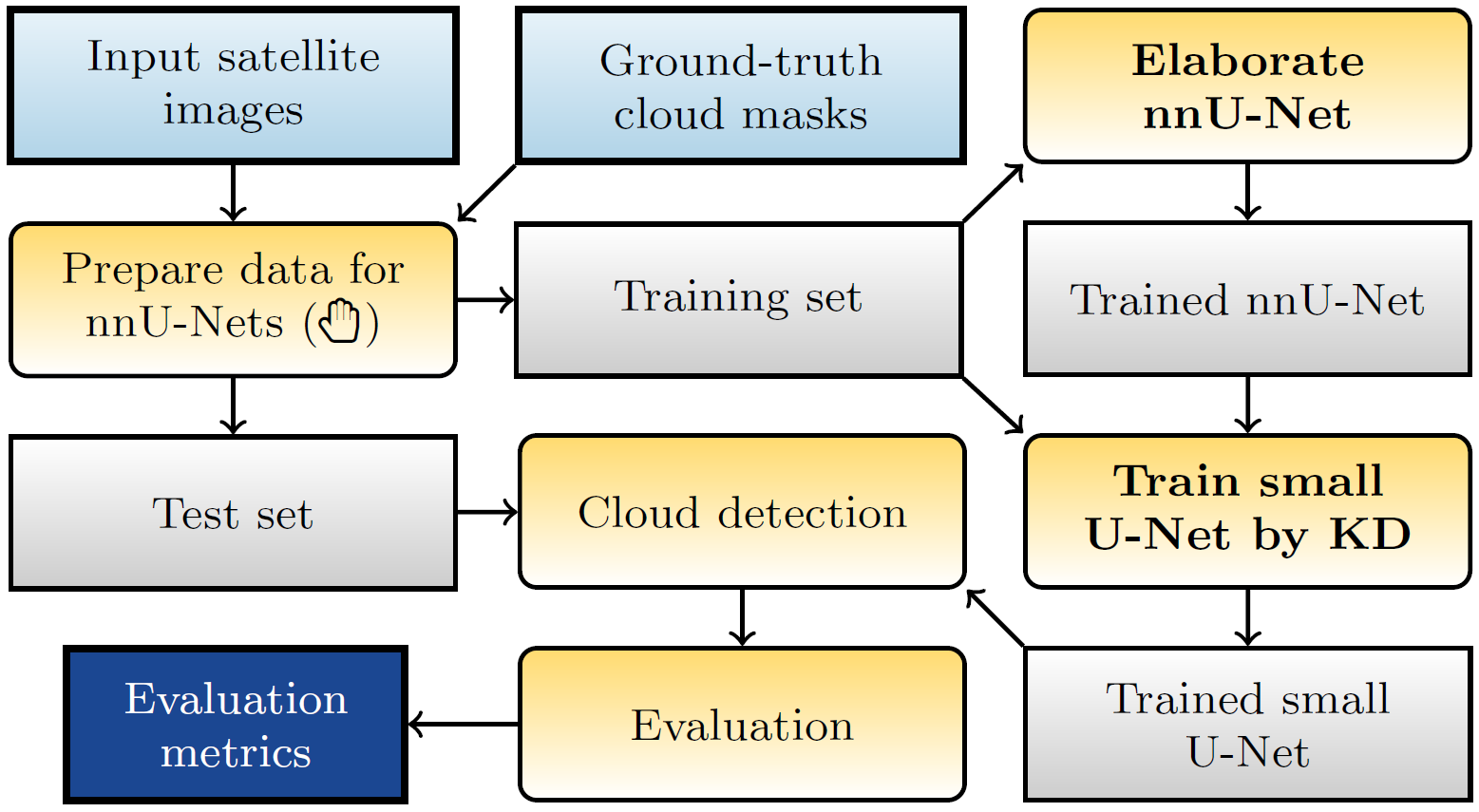}
     \caption{A high-level flowchart presenting an end-to-end deployment pipeline which exploits nnU-Nets. In bold, we present the steps which lead to generating fully-functional models for cloud detection.}
     \label{fig:nnunet_flowchart}
 \end{figure}

The nnU-Net algorithm is a deep learning segmentation method that automatically adapts itself based on the underlying characteristics of the training data and target segmentation problem. This configuration encompasses data pre-processing, network architecture, training parameters, basic post-processing, and ensembling of several U-Net-based models~\cite{isensee_nnu-net_2021}. In Figure~\ref{fig:nnunet_flowchart}, we present a high-level flowchart of a deployment pipeline for cloud detection using nnU-Nets. There is only one manual step (\faHandPaperO) which needs to be performed beforehand, and it involves preparing the image data, since nnU-Nets were originally developed for medical images.

Once the annotated training data is fed to the framework, it is transformed into a standardized dataset representation (it captures e.g.,~the image size or class imbalance ratio), and the nnU-Net re-configures itself to generate an end-to-end segmentation pipeline. The optimization is based on distilling the underlying domain knowledge into the rule-based and empirical hyper-parameters. Afterwards, a set of heuristics is used to model the dependencies between a selected parameter and the anticipated network's performance, e.g., ``larger patches increase the contextual information available, thus should be preferred''. The data-related knowledge is distilled into such rules by e.g.,\,initializing the initial patch size to the median image size within the training data, and iteratively decreasing it until the deep network can be trained with some GPU constraints\footnote{The detailed description of the nnU-Net adaptation rules and heuristics is included in the supplementary material of the work by Isensee et al.~\cite{isensee_nnu-net_2021}.}. The hyper-parameters that undergo adaptation encompass, among others, the network topology, patch and batch sizes, normalization, or building an ensemble of base nnU-Net models (including 2D, 3D, and cascaded 3D U-Net-based architectures). Although Isensee et al.~\cite{isensee_nnu-net_2021} claimed that the nnU-Net framework can be utilized in the biomedical domain and indeed showed its superior performance in more than 20 segmentation problems, this approach---to the best of our knowledge---has never been exploited to tackle the Earth observation image analysis tasks. We address this research gap and hypothesize that nnU-Nets can be deployed in a hands-free manner in the satellite image analysis, and that they can deliver competitive performance without any manual intervention. 

\begin{figure}[ht!]
    \centering
    \includegraphics[width=\columnwidth]{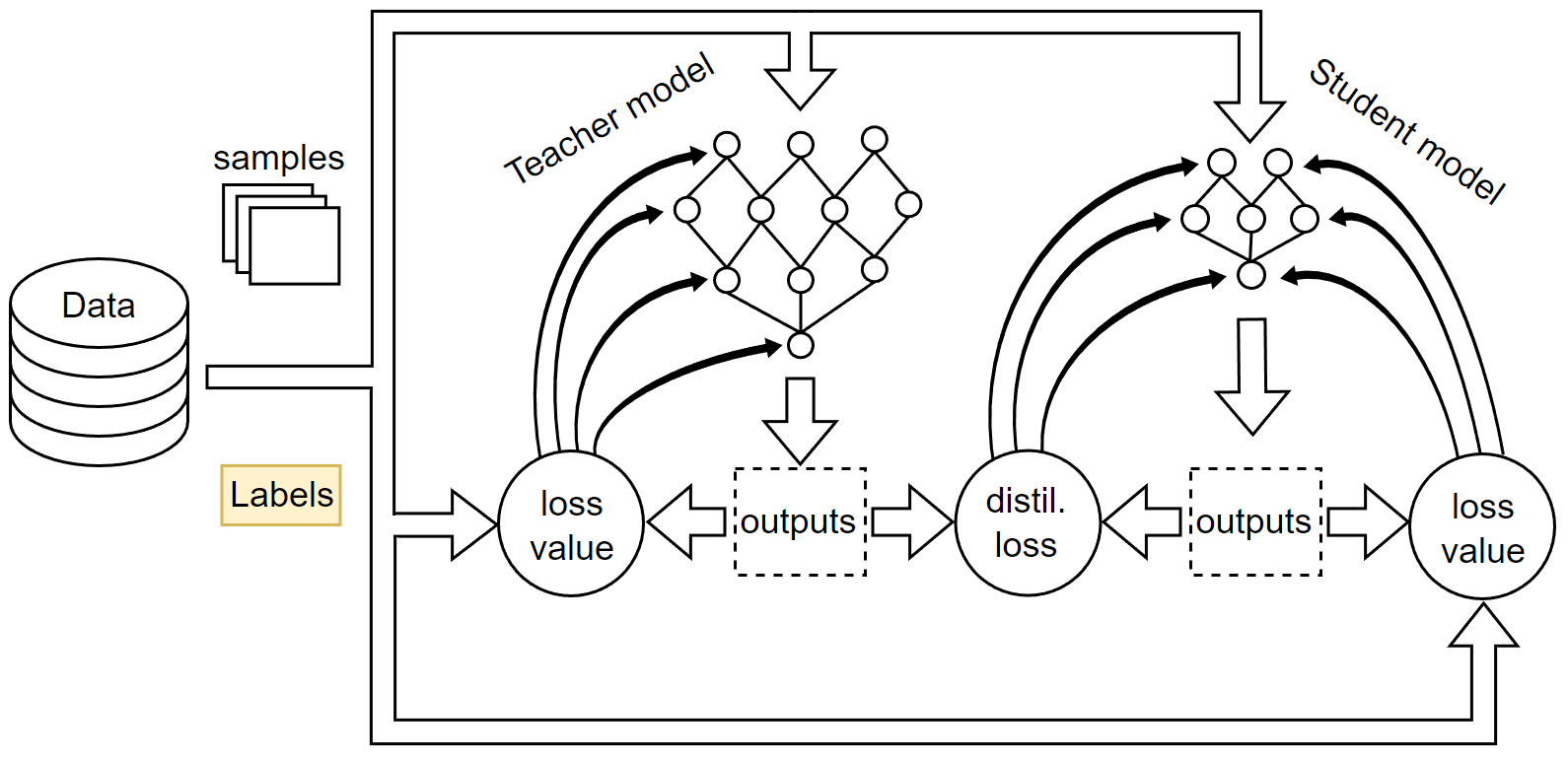}
    \caption{A visualization of the basic case of knowledge distillation. The student model is influenced by two factors: regular loss and distillation loss.}
    \label{distillation-scheme}
\end{figure}

Since nnU-Nets are very large in terms of the number of trainable parameters (easily reaching hundreds of millions), especially if several base nnU-Nets are ensembled together, they would not fit into hardware-constraints environments for on-board processing. To address this issue, we exploit knowledge distillation to train significantly smaller U-Nets (a student network, with hundreds of thousands of parameters) while still benefiting from not only available ground truth, but also from the characteristics of a large model (the teacher; see a schematic view of the KD process in Figure~\ref{distillation-scheme}). To make a student mimic the teacher's responses, the loss function is extended with a term for penalizing the difference between the student's and teacher's predictions~\cite{https://doi.org/10.48550/arxiv.1503.02531}:
\begin{dmath}
\mathcal{L}_s(x, \textbf{W}) = \alpha \mathcal{E}(y,\sigma(z_s,\tau_0))\\+(1-\alpha)\mathcal{E}(\sigma(z_t,\tau), \sigma(z_s.\tau)),
\end{dmath}
where \textbf{W} are the student model parameters, $y$ is the GT label, $\mathcal{E}$ is a cross-entropy loss function, and $z_s$ and $z_t$ are soft responses of the student and the teacher, $\sigma$ is a softmax function parameterized by the temperature $\tau$ ($\tau_0=1$), $\alpha \in (0, 1)$ balances the impact of self-teaching and the teacher (we use the parameterization as suggested in~\cite{https://doi.org/10.48550/arxiv.1503.02531}). Although nnU-Nets can be effectively distilled into different architectures, we focus on an optimized U-Net as---albeit only 450k parameters---it was shown robust against different in-orbit conditions, and they are easy to implement using Xilinx tools, commonly used in AI-powered space missions~\cite{9554170}.

\section{Experimental Results}\label{sec:experiments}

\newcommand{\Jaccard}{JI}
\newcommand{\Precision}{Pr}
\newcommand{\Recall}{Re}
\newcommand{\Specificity}{Spe}
\newcommand{\OverallAccuracy}{OA}

The objectives of the experimental study are three-fold: to (\textit{i})~confront the nnU-Nets for cloud detection with other state-of-the-art techniques, (\textit{ii})~to verify their flexibility over satellite imagery captured using different missions, and (\textit{iii})~investigate compact U-Nets trained from scratch and benefiting from large-capacity nnU-Nets through KD. To quantify the performance of our models over the 38-C dataset, we use the Jaccard index (\Jaccard), precision (\Precision), recall (\Recall), specificity (\Specificity) and overall accuracy (\OverallAccuracy), whereas for the OCN we report JI only, as only this metric was calculated by the independent validation server (the same metric is used for emulated $\Phi$-Sat-2 data in the KD experiment). All results are reported for the test sets (unless stated otherwise) that were unseen during the training process which lasted for 1000 epochs. The experiments ran on an NVIDIA RTX 3090 GPU with 24 GB VRAM, and a single training process took approx. 48 h. The resulting nnU-Net models, together with their entire parameterization are available at \url{https://gitlab.com/jnalepa/nnUNets_for_clouds}.

\begin{table}[ht!]
\centering
\scriptsize
  \caption{Comparison of the nnU-Nets with other techniques specifically designed to detect clouds in L-8 imagery. The best metrics are boldfaced, whereas the second best are underlined.}
  \label{tab:results}
  \renewcommand{\tabcolsep}{3.9mm}
  \begin{tabular}{rcccccc}
    \Xhline{2\arrayrulewidth}
    Algorithm & \Jaccard & \Precision & \Recall & \Specificity & \OverallAccuracy\\
    \hline
    FCN$\dagger$~\cite{38-cloud-2} & 0.722 & \uline{0.846} & 0.814 & \uline{0.985} & 0.952\\
    Fmask~\cite{ZHU2015269} & 0.752 & 0.777 & \textbf{0.972} & 0.940 & 0.949\\
    Cloud-Net~\cite{38-cloud-1} & \textbf{0.785} & \textbf{0.912} & 0.849 & \textbf{0.987} & \textbf{0.965}\\
    nnU-Net & \uline{0.756} & {0.845} & \uline{0.866} & 0.976 & \uline{0.953}\\
    \Xhline{2\arrayrulewidth}
    \multicolumn{6}{l}{$\dagger$ Trained with the 38-C training dataset.}
\end{tabular}
\end{table}

\begin{figure}[ht!]
\centering
\includegraphics[width=0.87\columnwidth]{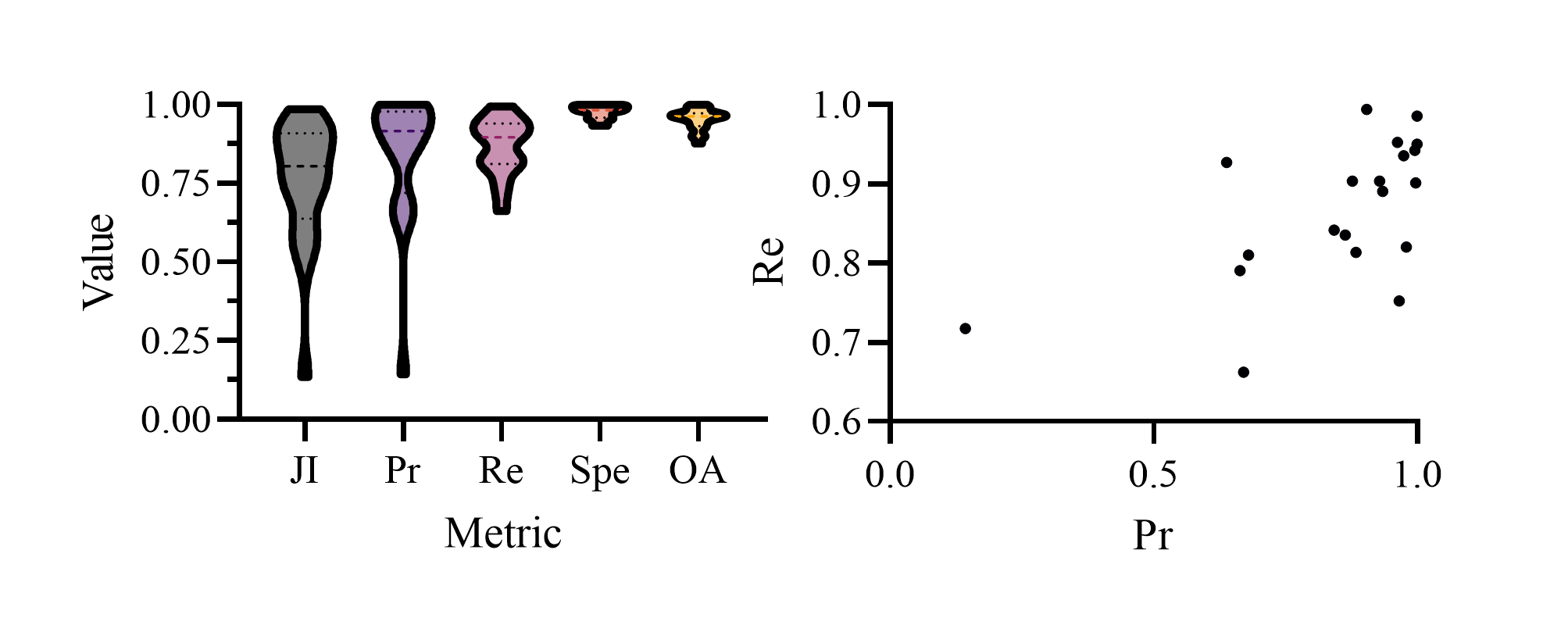}\\[-0.5cm]
\caption{The distribution of all metrics obtained using the nnU-Nets across all test 38-C scenes (left), together with the corresponding precision-recall scatter plot (right). The most challenging scene (\Precision: 0.143 and \Recall: 0.717) presents the mountainous areas for which the nnU-Nets delivered false-positive detections. It is of note that the L-8 test set contains the snowy areas which are challenging to segment due to a large number of bright pixels that can be mislabeled as clouds (leading to a significant number of false positives).}\label{fig:38c_violin}
\end{figure}

\begin{figure}[ht!]
\centering
\scriptsize
\renewcommand{\tabcolsep}{0.2mm}
\newcommand{\mymagnification}{6}
\newcommand{\mywidth}{0.29}
\begin{tabular}{cccc}
\Xhline{2\arrayrulewidth}
     & RGB & Ground truth & nnU-Net\\
     \hline
     \rotatebox{90}{~Best (\Jaccard: 0.985)} & \begin{tikzpicture}
        [,spy using outlines={circle,pink,magnification=\mymagnification,size=1.5cm, connect spies}]
        \node {\pgfimage[width=\mywidth\columnwidth]{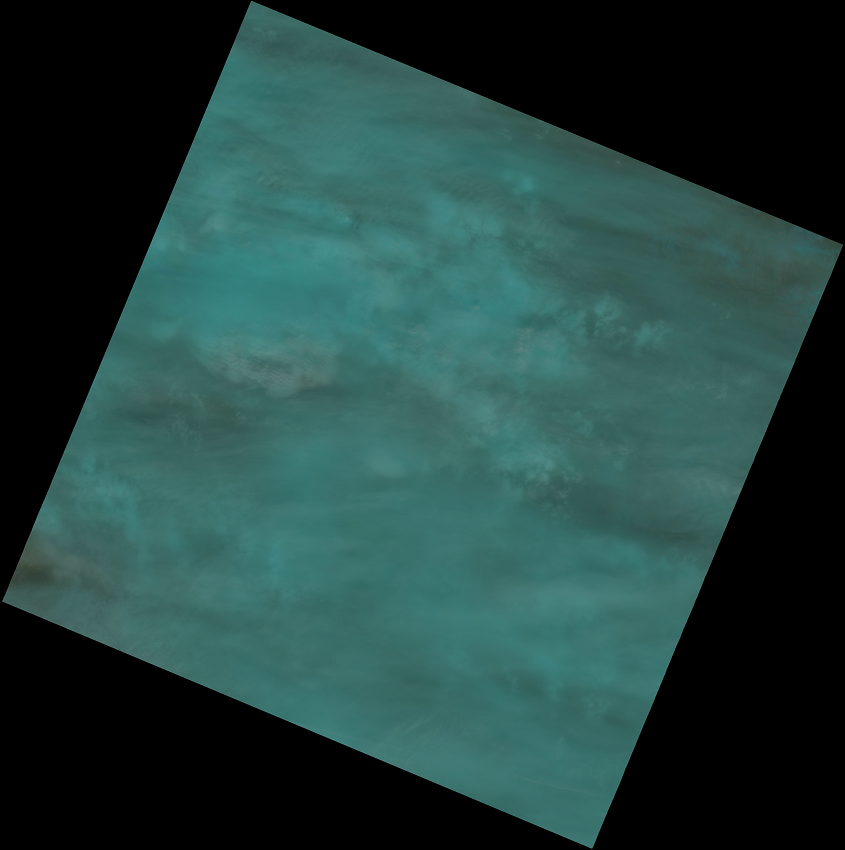}};
        \spy[every spy on node/.append style={thick},every spy in node/.append style={thick}] on (1.1,0.527) in node [left] at (1.1,-0.5);
        \end{tikzpicture}& \begin{tikzpicture}
        [,spy using outlines={circle,pink,magnification=\mymagnification,size=1.5cm, connect spies}]
        \node {\pgfimage[width=\mywidth\columnwidth]{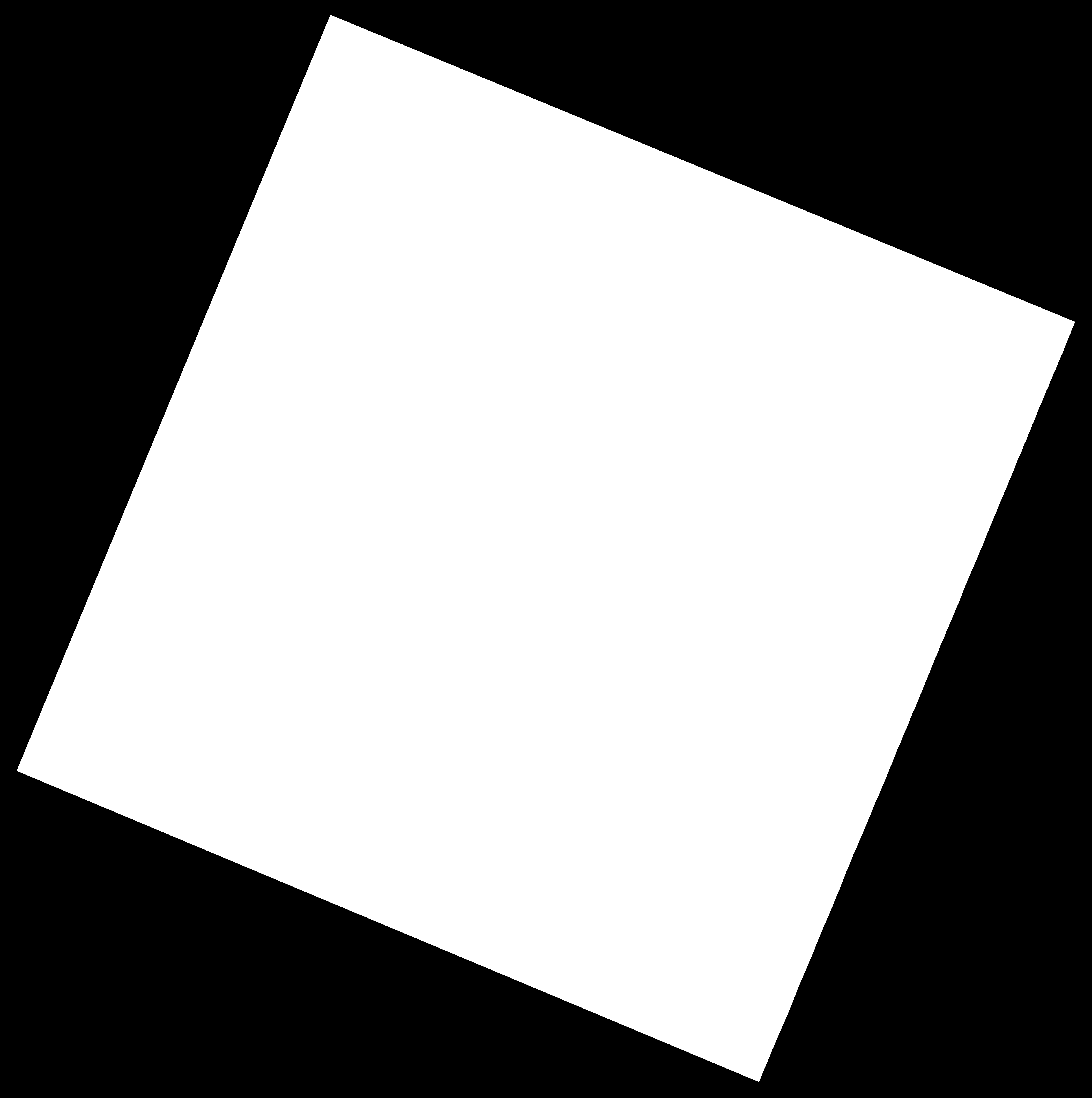}};
        \spy[every spy on node/.append style={thick},every spy in node/.append style={thick}] on (1.1,0.5) in node [left] at (1.1,-0.5);
        \end{tikzpicture}& \begin{tikzpicture}
        [,spy using outlines={circle,pink,magnification=\mymagnification,size=1.5cm, connect spies}]
        \node {\pgfimage[width=\mywidth\columnwidth]{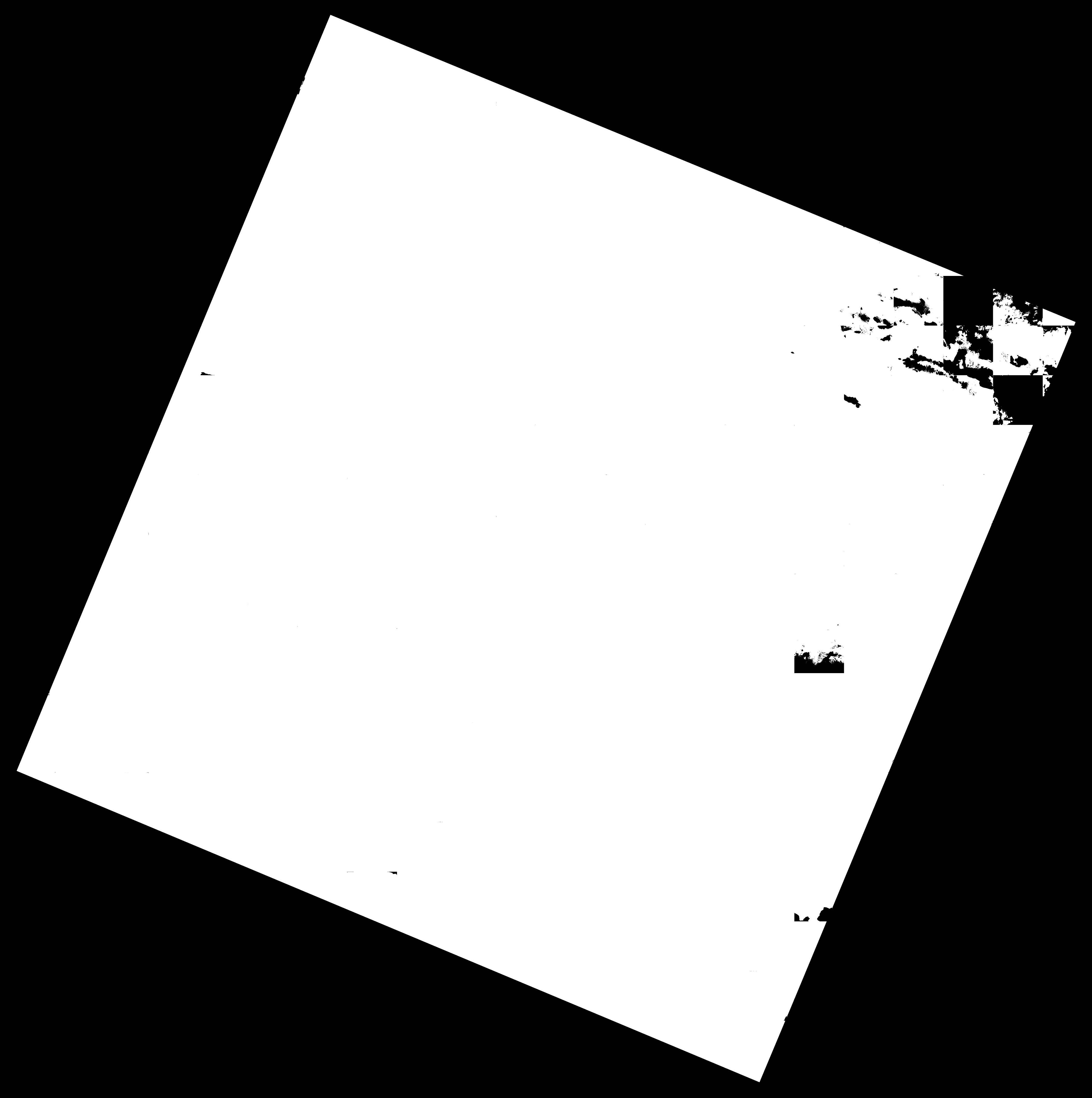}};
        \spy[every spy on node/.append style={thick},every spy in node/.append style={thick}] on (1.1,0.5) in node [left] at (1.1,-0.5);
        \end{tikzpicture}\\[-0.1cm]
     \hline
     \rotatebox{90}{~Median (\Jaccard: 0.806)} &\begin{tikzpicture}
        [,spy using outlines={circle,pink,magnification=\mymagnification,size=1.5cm, connect spies}]
        \node {\pgfimage[width=\mywidth\columnwidth]{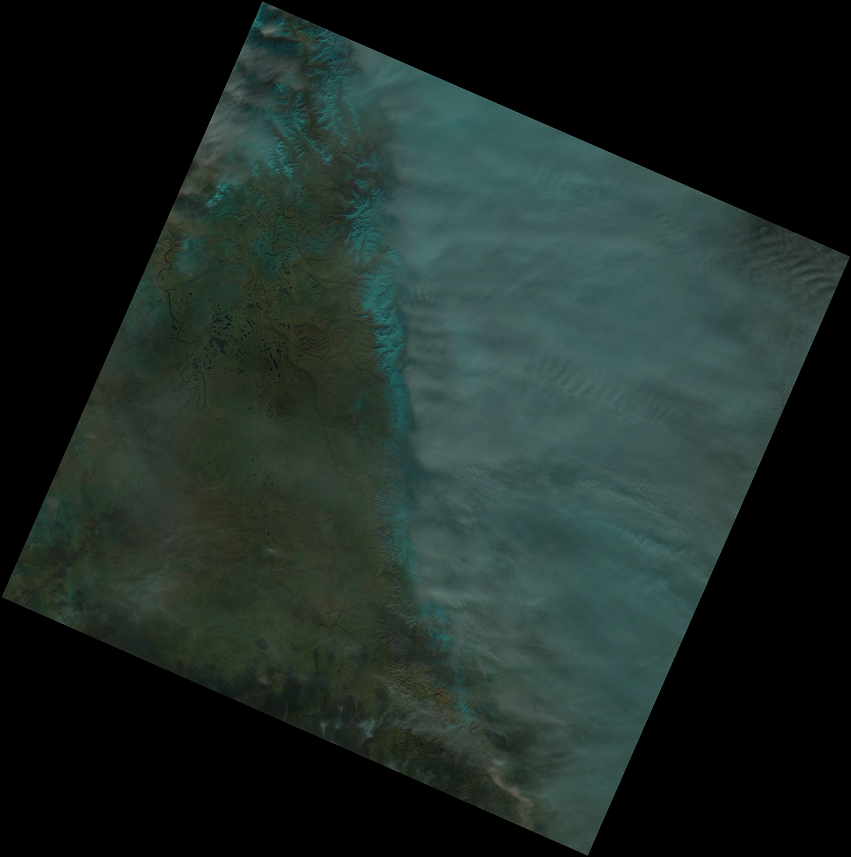}};
        \spy[every spy on node/.append style={thick},every spy in node/.append style={thick}] on (-0.4,0.5) in node [left] at (1.1,-0.5);
        \end{tikzpicture}& \begin{tikzpicture}
        [,spy using outlines={circle,pink,magnification=\mymagnification,size=1.5cm, connect spies}]
        \node {\pgfimage[width=\mywidth\columnwidth]{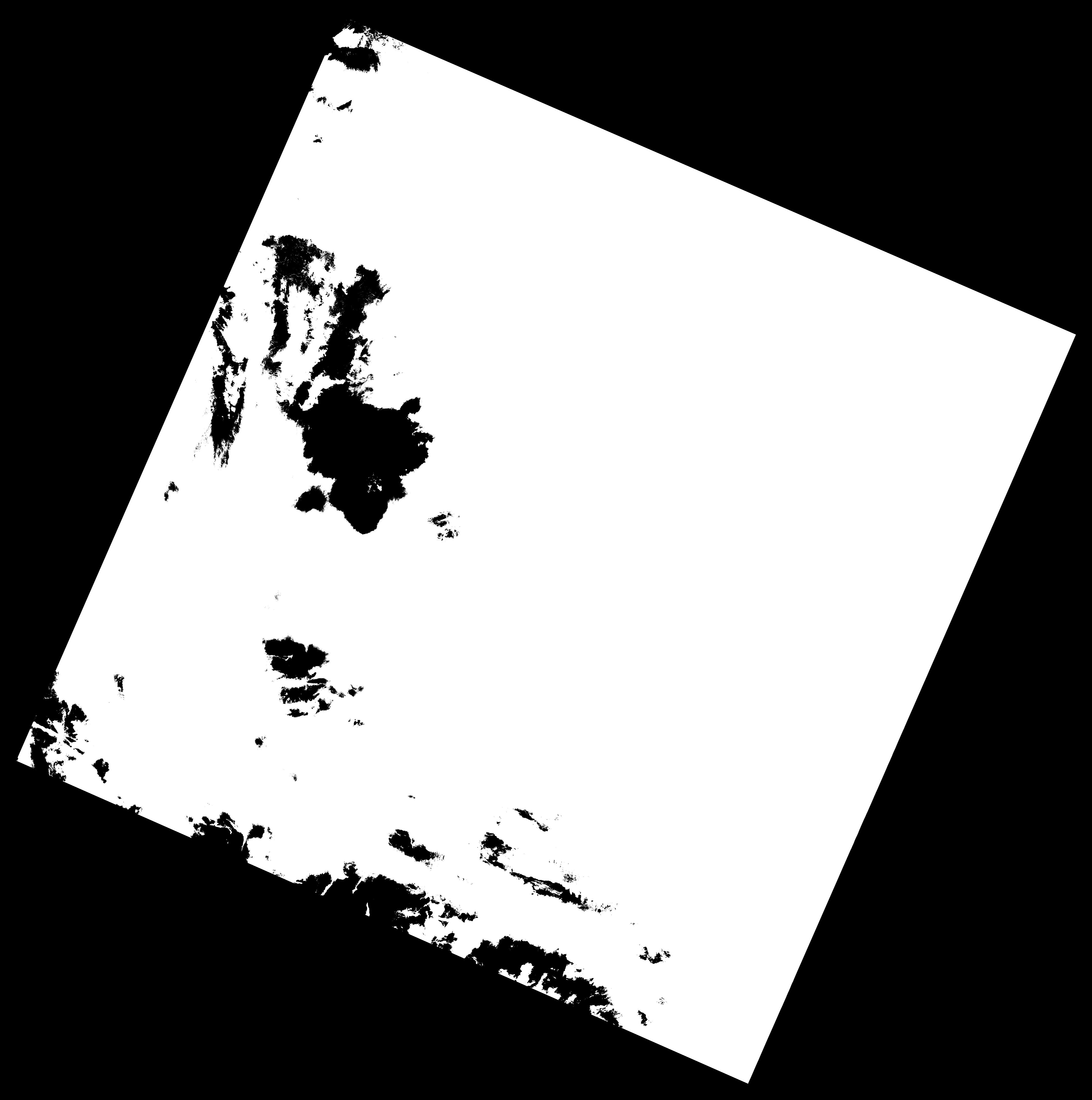}};
        \spy[every spy on node/.append style={thick},every spy in node/.append style={thick}] on (-0.4,0.5) in node [left] at (1.1,-0.5);
        \end{tikzpicture}& \begin{tikzpicture}
        [,spy using outlines={circle,pink,magnification=\mymagnification,size=1.5cm, connect spies}]
        \node {\pgfimage[width=\mywidth\columnwidth]{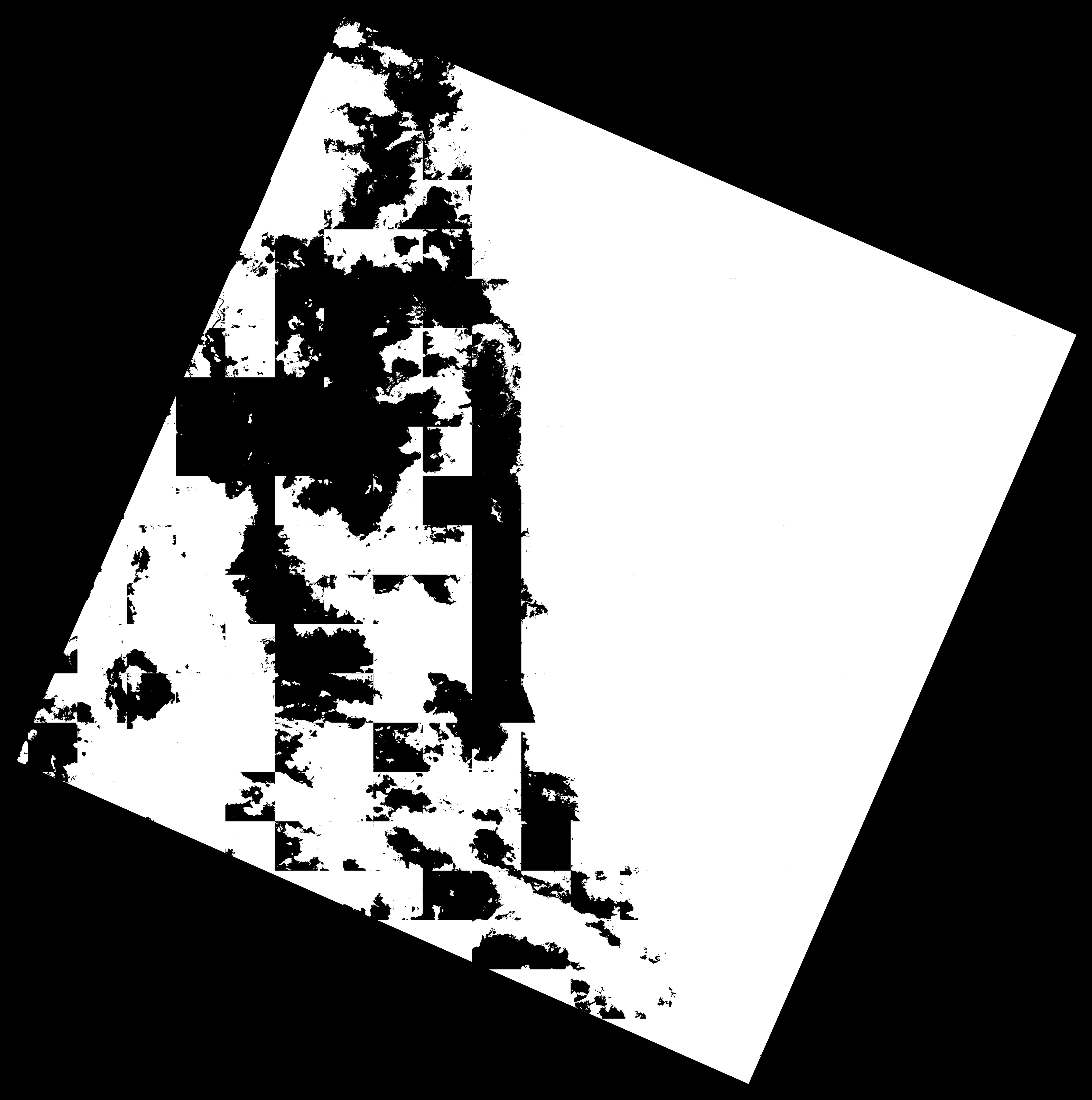}};
        \spy[every spy on node/.append style={thick},every spy in node/.append style={thick}] on (-0.4,0.5) in node [left] at (1.1,-0.5);
        \end{tikzpicture}\\[-0.1cm]
     \hline
     \rotatebox{90}{~Worst (\Jaccard: 0.135)} &\begin{tikzpicture}
        [,spy using outlines={circle,pink,magnification=\mymagnification,size=1.5cm, connect spies}]
        \node {\pgfimage[width=\mywidth\columnwidth]{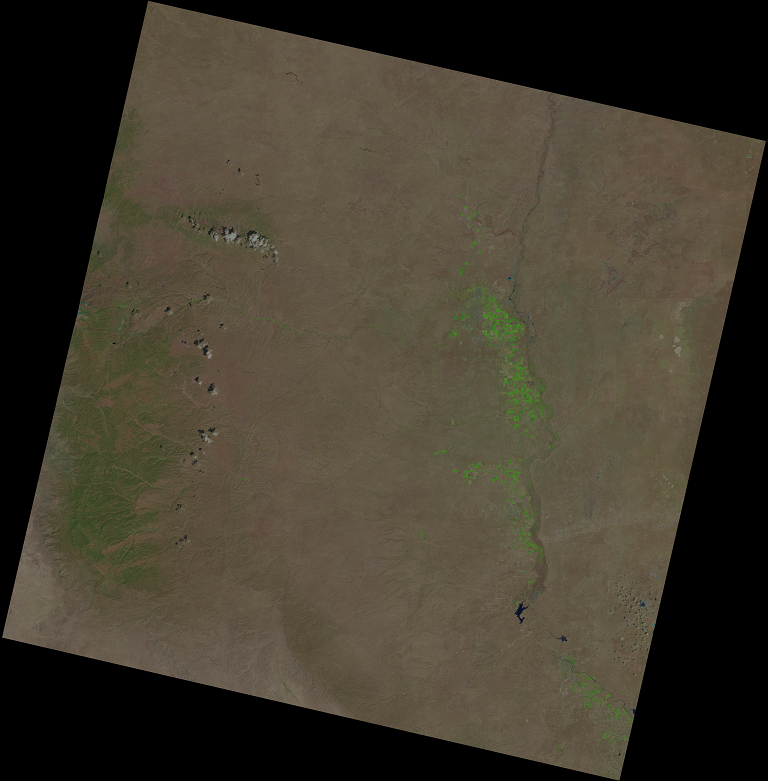}};
        \spy[every spy on node/.append style={thick},every spy in node/.append style={thick}] on (-1.15,-0.4) in node [left] at (1.1,-0.5);
        \end{tikzpicture}&
     \begin{tikzpicture}
        [,spy using outlines={circle,pink,magnification=\mymagnification,size=1.5cm, connect spies}]
        \node {\pgfimage[width=\mywidth\columnwidth]{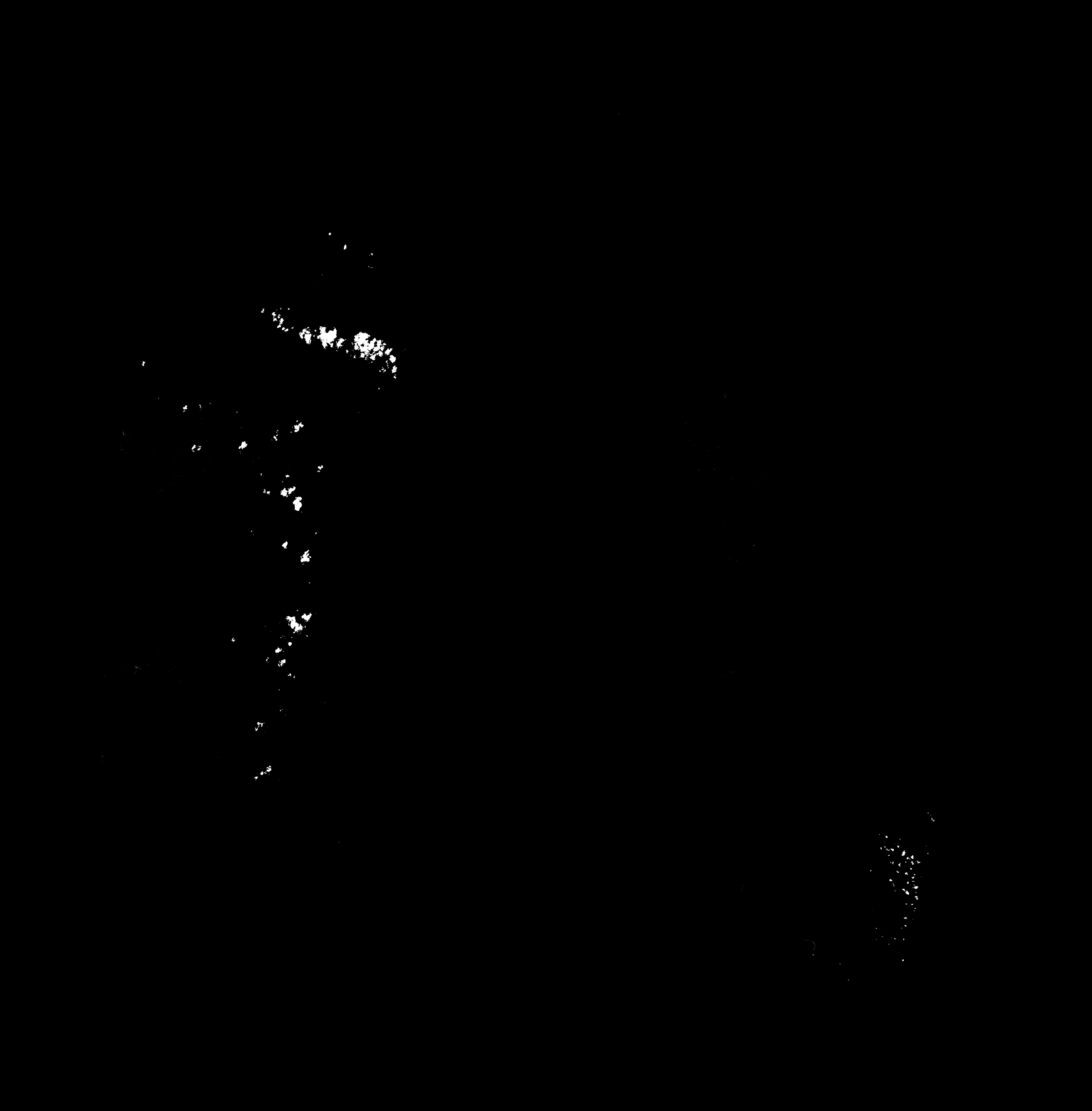}};
        \spy[every spy on node/.append style={thick},every spy in node/.append style={thick}] on (-1.15,-0.4) in node [left] at (1.1,-0.5);
        \end{tikzpicture}&
     \begin{tikzpicture}
        [,spy using outlines={circle,pink,magnification=\mymagnification,size=1.5cm, connect spies}]
        \node {\pgfimage[width=\mywidth\columnwidth]{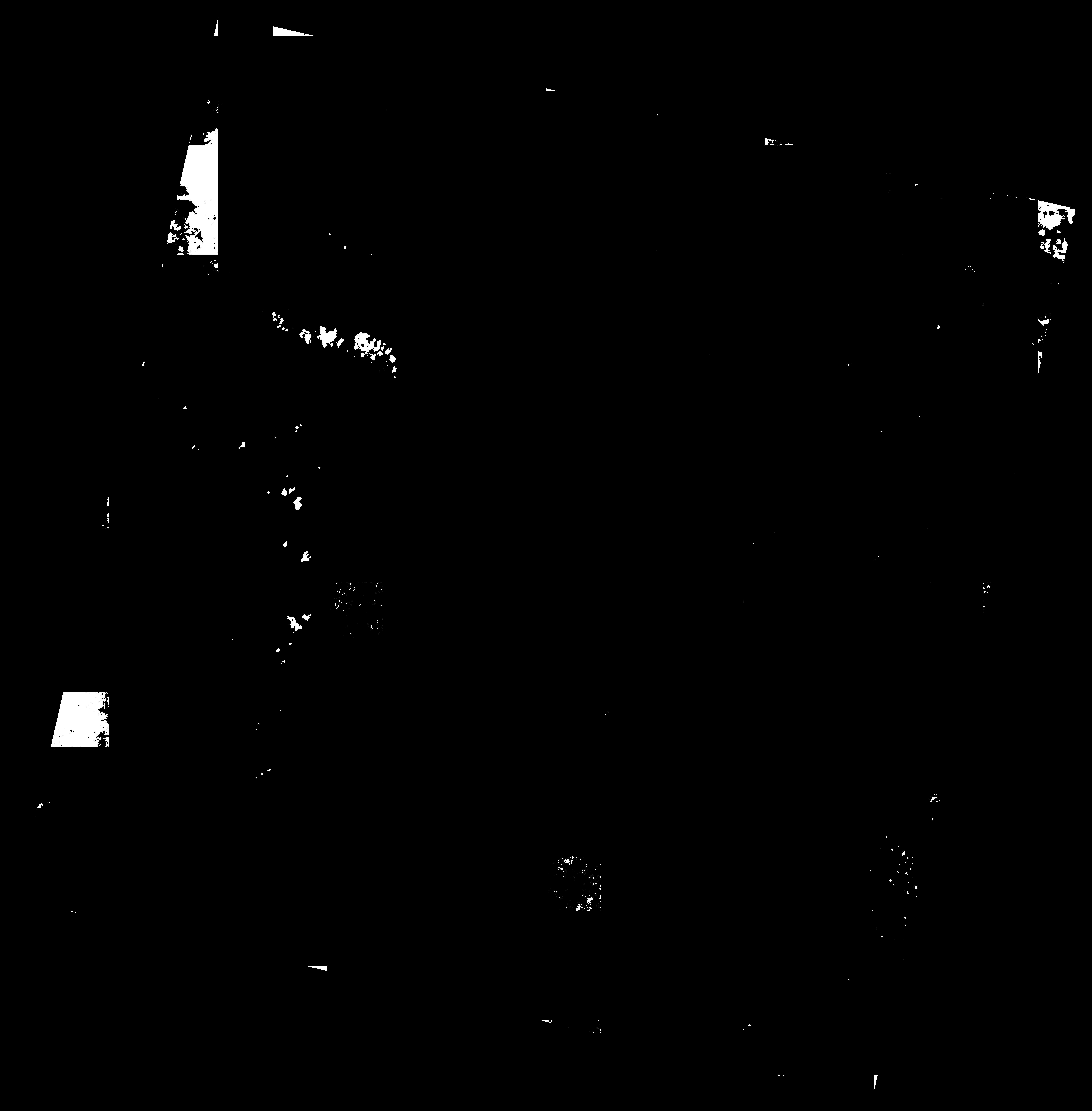}};
        \spy[every spy on node/.append style={thick},every spy in node/.append style={thick}] on (-1.15,-0.4) in node [left] at (1.1,-0.5);
        \end{tikzpicture}\\[-0.1cm]
     \Xhline{2\arrayrulewidth}
\end{tabular}
\caption{Example 38-C cloud masks obtained using the nnU-Nets.}\label{fig:examples}
\end{figure}

The results obtained for the test 38-C patches gathered in Table~\ref{tab:results} indicate that the ensemble of five 3D nnU-Nets (with approximately 25M trainable parameters; processing all 10k test patches took less than one hour) offers competitive performance which is on par with the performance delivered by the methods specifically designed to detect clouds in the L-8 multispectral images. We compared the proposed method with the improved Fmask algorithm that was fine-tuned for L-8 images~\cite{ZHU2015269}, Cloud-Net~\cite{38-cloud-1}, and a more generic fully-convolutional U-Net network trained over the training 38-C data (FCN)~\cite{38-cloud-2}. Although we cannot claim that the nnU-Nets outperformed the hand-crafted techniques, we emphasize that the nnU-Nets were obtained without any manual design procedures or user intervention. The distribution of the metrics (Figure~\ref{fig:38c_violin}) shows that the nnU-Nets extracted high-quality cloud masks for the vast majority of scenes. In Figure~\ref{fig:examples}, we additionally present the visual examples: best, worst, and median-quality segmentation as measured by \Jaccard).

To investigate the flexibility of the nnU-Net pipeline, we exploited it for the S-2 data within the On Cloud N: Cloud Cover Detection Challenge. Here, due to the execution time constraints imposed by the organizers (4\,h for processing the entire test set of more than 10k S-2 patches), we exploited an ensemble of four (instead of five) 3D nnU-Nets (approx. 30M trainable parameters). In this experiment, we additionally employed a manually-designed heuristic post processing routine based on the morphological alterations of the resulting nnU-Net cloud masks. In the nnU-Net with our post processing, we execute the closing operation (with the $3\times 3$ kernel) if more than 50\% of pixels within the patch are annotated as clouds in the segmentation map (otherwise, we apply the opening operation with the same kernel). This approach was hand-crafted based on the visual inspection of the results obtained for both training and test S-2 patches (for the latter, we did not have the GT). We observed that there were patches with small false-positive objects (especially in the snowy areas), whereas for the cloudy scenes, we noticed under-segmentation.

\begin{figure}[ht!]
\centering
\includegraphics[width=0.9\columnwidth]{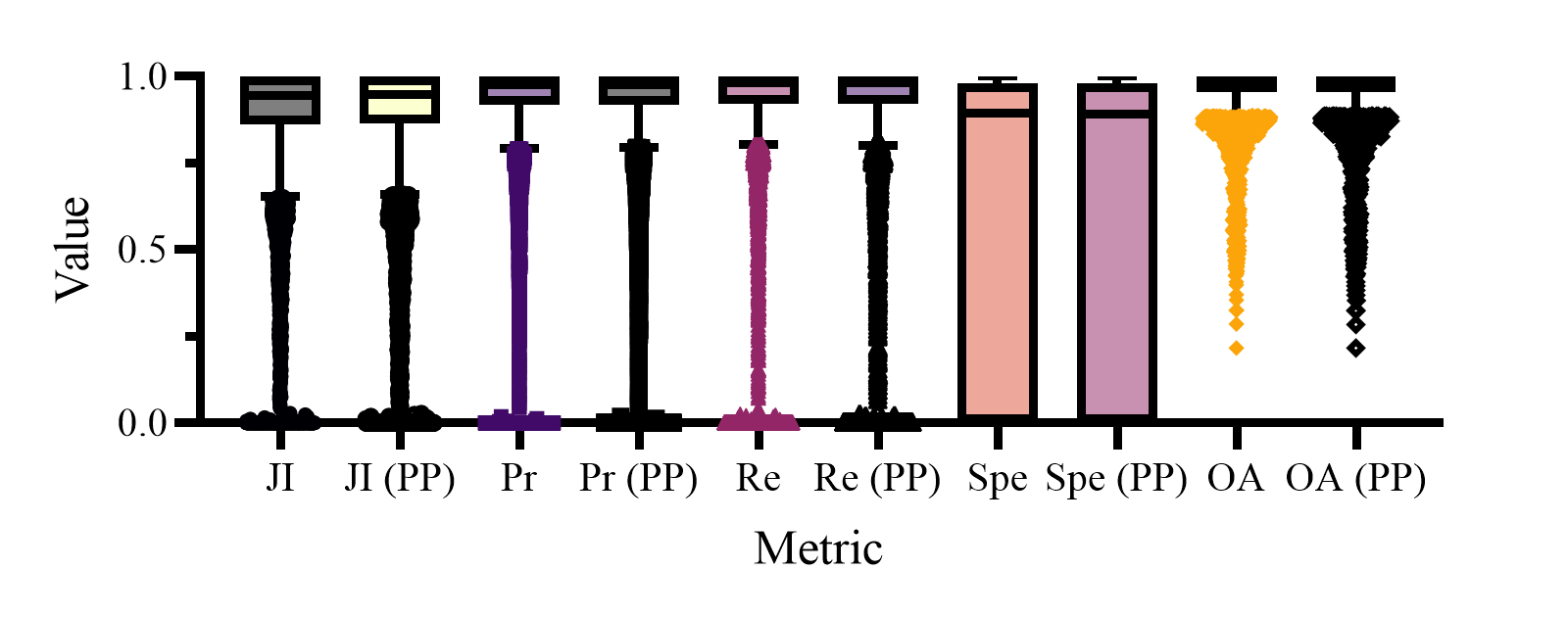}\\[-0.3cm]
\caption{The Tukey box plots showing the distribution of the metrics obtained for approx. 12k \textit{training} S-2 patches from the OCN set using the nnU-Nets with (PP) and without manually-designed post processing. The individual points present outliers (values lower than the 25$^{\rm th}$ percentile of the metric minus 1.5 inter-quartile distance). }\label{fig:s2_tukey}
\end{figure}

Although the differences (Figure~\ref{fig:s2_tukey}) between the nnU-Net models with and without post processing are statistically significant in the pairwise comparisons over the training data (Wilcoxon two-tailed tests, $p<0.05$), those improvements only marginally affect the scores obtained over the test data (\Jaccard\, increased from 0.8822 to 0.8824). It indicates that the nnU-Net framework automatically elaborates well-performing models which are extremely challenging to improve manually. Overall, we have investigated 18 variants of the nnU-Nets with various post processing, updated loss functions, and without and with ensembling (in the former approach, we trained a single 2D/3D model over the entire training set, whereas in the latter, four deep models are trained independently over non-overlapping folds, and then are fused within the nnU-Net), and none of them delivered further improvements in \Jaccard.

\begin{table}[ht!]
\centering
\scriptsize
  \caption{Comparison of the nnU-Nets with other techniques within the On Cloud N: Cloud Cover Detection Challenge. The result (average \Jaccard) obtained by the nnU-Nets is boldfaced.}
  \label{tab:results_challenge}
  \renewcommand{\tabcolsep}{1mm}
  \begin{tabular}{rcccccccccc}
    \Xhline{2\arrayrulewidth}
    Rank & 1 & 2 & 3 & 10 & 20 & 30 & 40 & \textbf{53} & 107 & S-2\\
    \hline
    \Jaccard & 0.897 & 0.897 & 0.896 & 0.895 & 0.894 & 0.892 & 0.889 & \textbf{0.882} & 0.815 & 0.652\\
    \Xhline{2\arrayrulewidth}
    \multicolumn{11}{l}{1$^{\rm st}$: Manually-designed ensemble of U-Nets with various backbones.}\\
    \multicolumn{11}{l}{2$^{\rm nd}$: U-Net++ models with manually-reviewed training data.}\\
    \multicolumn{11}{l}{3$^{\rm rd}$: Manually-designed ensemble of U-Nets with different pre-trained encoders.}\\
    \multicolumn{11}{l}{107$^{\rm th}$: Baseline model (U-Net with pre-trained ResNet-34 used as the encoder).}\\
    \multicolumn{11}{l}{S-2: Built-in S-2 cloud detection (thresholding of the cirrus B10 band).}
\end{tabular}
\end{table}

In Table~\ref{tab:results_challenge}, we gather the results obtained by the participating teams over the test S-2 patches (note that other participants might have used other S-2 bands too). We can observe that the differences between our nnU-Nets (with manual post processing applied) and the top 3 teams amount to 0.015, 0.015, and 0.014, respectively. It is of note that all the top-ranked techniques were manually-crafted to this challenge (and underlying data), either at the architectural level (rank 1 and 3), or also at the data level (rank 2), where the authors did review the training patches in a semi-automated way to remove the patches for which the GT quality was questionable. On the other hand, the difference in \Jaccard\, is indeed more visible between the baseline model (ranked 107) and all of the aforementioned ones---it was 0.082 and 0.067 for the winners and our solution.

To verify if small differences (for 38-C, the difference in \Jaccard\, between Cloud-Net and the nnU-Nets was 0.029, whereas for the OCN test data it was 0.015 between the winners and our approach) are possible to spot by the naked eye, we conducted the mean opinion score (MOS) experiment. The responders were given 15 images from the training OCN data with two cloud masks obtained using the nnU-Nets with and without post processing, and the task was to select the mask which ``\emph{more precisely presents the clouds}''. The participants could say that ``\emph{both masks look equally good to me}'' and ``\emph{none}'' (if neither mask was good enough based on the visual inspection). The masks differed in quality measured by \Jaccard\, (min., avg., median, and max. difference in \Jaccard\, was 0.020, 0.030, 0.027, and 0.047). The nnU-Nets with post processing gave the better \Jaccard\,in 6/15 cases (with the avg. improvement in \Jaccard\, of 0.031), whereas in the remaining 9/15 cases the nnU-Nets without post processing resulted in \Jaccard\, better by 0.029 on average. The background of the participants (109 in total) was diverse---we announced MOS at KP Labs, ESA, but also at the university across the students with no remote sensing experience. 

\begin{table}[ht!]
\centering
\scriptsize
  \caption{Average percentage of MOS responses indicating that mask A/B (without/with post processing) was selected as better, and avg. percentage of MOS responses indicating that both masks were equally good or none was good enough according to the responders.}
  \label{tab:mos}
  \renewcommand{\tabcolsep}{2mm}
  \begin{tabular}{cccccc}
    \Xhline{2\arrayrulewidth}
    Images & Mask A (avg. \Jaccard) & Mask B (avg. \Jaccard) & Both & None\\
    \hline
    Higher \Jaccard\, for Mask A & 34.86 (0.485)	& 12.11	(0.456) & 25.32	& 27.73\\
    Higher \Jaccard\, for Mask B & 28.05 (0.698) & 26.58 (0.729)	& 12.87	& 32.52\\
    \hline
    All images & 32.13 (0.570) & 17.90 (0.565) & 20.34	& 29.65\\
    \Xhline{2\arrayrulewidth}
\end{tabular}
\end{table}

\begin{figure}[ht!]
\centering
\scriptsize
\renewcommand{\tabcolsep}{0.2mm}
\newcommand{\mymagnification}{6}
\newcommand{\mywidth}{0.29}
\begin{tabular}{cccc}
\Xhline{2\arrayrulewidth}
     & a) None ($\Delta$: 0.047) & b) Without PP ($\Delta$: 0.021) & c) With PP ($\Delta$: 0.022)\\
     \hline
     \rotatebox{90}{~RGB} & \begin{tikzpicture}
        [,spy using outlines={circle,pink,magnification=\mymagnification,size=1.5cm, connect spies}]
        \node {\pgfimage[width=\mywidth\columnwidth]{"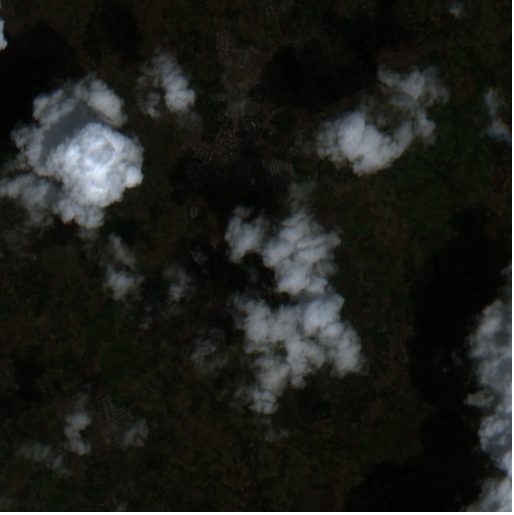"}};
        \end{tikzpicture}& 
        \begin{tikzpicture}
        [,spy using outlines={circle,pink,magnification=\mymagnification,size=1.5cm, connect spies}]
        \node {\pgfimage[width=\mywidth\columnwidth]{"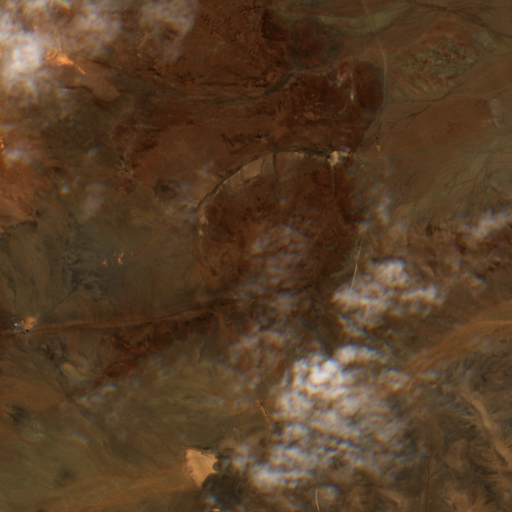"}};
        \end{tikzpicture}& 
        \begin{tikzpicture}
        [,spy using outlines={circle,pink,magnification=\mymagnification,size=1.5cm, connect spies}]
        \node {\pgfimage[width=\mywidth\columnwidth]{"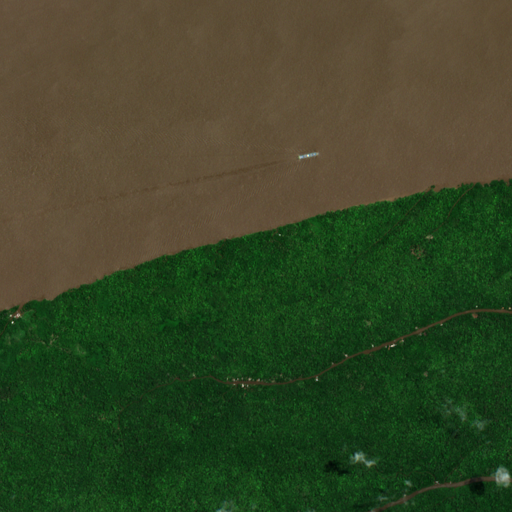"}};
        \spy[every spy on node/.append style={thick},every spy in node/.append style={thick}] on (0.25,0.5) in node [left] at (0.22,-0.5);
        \end{tikzpicture}\\[-0.1cm]
     \hline
     \rotatebox{90}{~Ground truth} &\begin{tikzpicture}
        [,spy using outlines={circle,pink,magnification=\mymagnification,size=1.5cm, connect spies}]
        \node {\pgfimage[width=\mywidth\columnwidth]{"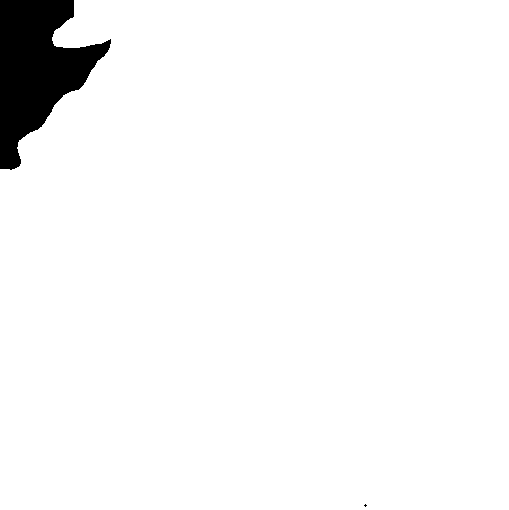"}};
        \end{tikzpicture}& \begin{tikzpicture}
        [,spy using outlines={circle,pink,magnification=\mymagnification,size=1.5cm, connect spies}]
        \node {\pgfimage[width=\mywidth\columnwidth]{"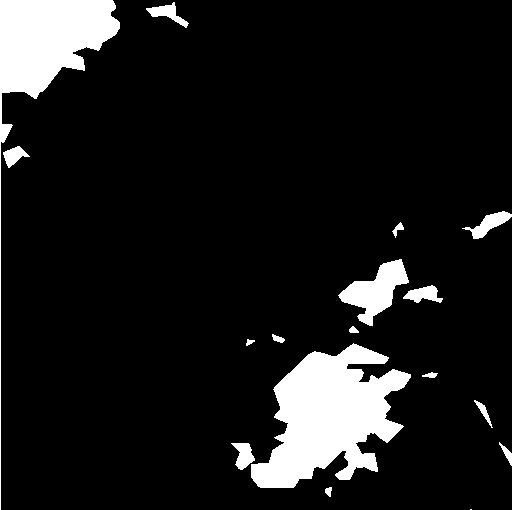"}};
        \end{tikzpicture}& \begin{tikzpicture}
        [,spy using outlines={circle,pink,magnification=\mymagnification,size=1.5cm, connect spies}]
        \node {\pgfimage[width=\mywidth\columnwidth]{"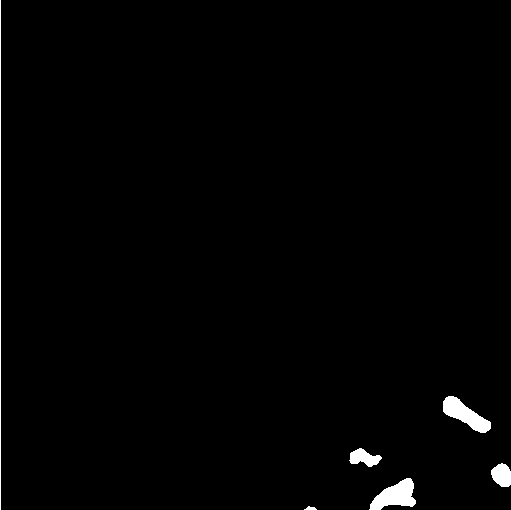"}};
        \spy[every spy on node/.append style={thick},every spy in node/.append style={thick}] on (0.25,0.5) in node [left] at (0.22,-0.5);
        \end{tikzpicture}\\[-0.1cm]
     \hline
     \rotatebox{90}{~Without PP} &\begin{tikzpicture}
        [,spy using outlines={circle,pink,magnification=\mymagnification,size=1.5cm, connect spies}]
        \node {\pgfimage[width=\mywidth\columnwidth]{"S2_examples_12_no"}};
        \end{tikzpicture}&
     \begin{tikzpicture}
        [,spy using outlines={circle,pink,magnification=\mymagnification,size=1.5cm, connect spies}]
        \node {\pgfimage[width=\mywidth\columnwidth]{"S2_examples_6_no"}};
        \end{tikzpicture}&
     \begin{tikzpicture}
        [,spy using outlines={circle,pink,magnification=\mymagnification,size=1.5cm, connect spies}]
        \node {\pgfimage[width=\mywidth\columnwidth]{"S2_examples_9_no"}};
        \spy[every spy on node/.append style={thick},every spy in node/.append style={thick}] on (0.25,0.5) in node [left] at (0.22,-0.5);
        \end{tikzpicture}\\[-0.1cm]
        \hline
     \rotatebox{90}{~With PP} &\begin{tikzpicture}
        [,spy using outlines={circle,pink,magnification=\mymagnification,size=1.5cm, connect spies}]
        \node {\pgfimage[width=\mywidth\columnwidth]{"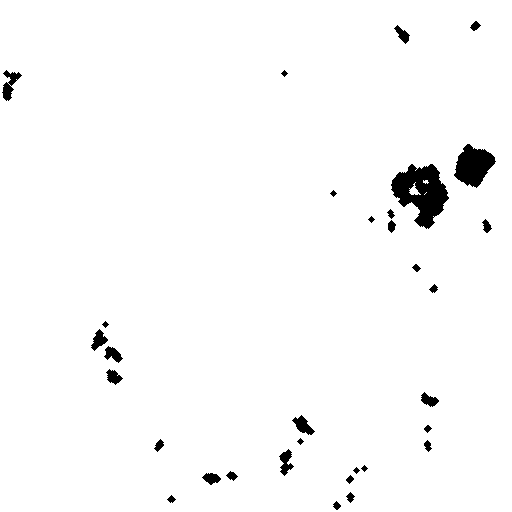"}};
        \end{tikzpicture}&
     \begin{tikzpicture}
        [,spy using outlines={circle,pink,magnification=\mymagnification,size=1.5cm, connect spies}]
        \node {\pgfimage[width=\mywidth\columnwidth]{"S2_examples_6_yes"}};
        \end{tikzpicture}&
     \begin{tikzpicture}
        [,spy using outlines={circle,pink,magnification=\mymagnification,size=1.5cm, connect spies}]
        \node {\pgfimage[width=\mywidth\columnwidth]{"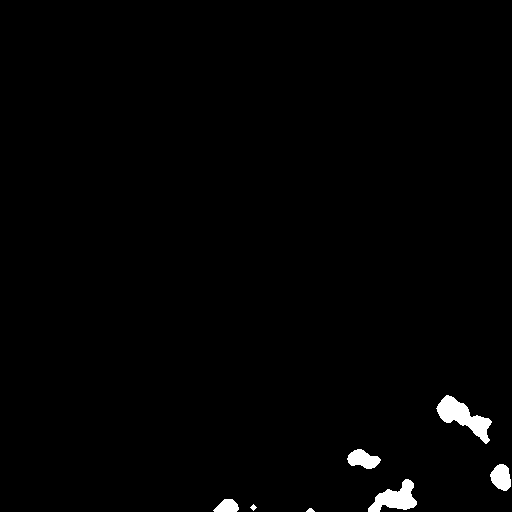"}};
        \spy[every spy on node/.append style={thick},every spy in node/.append style={thick}] on (0.25,0.5) in node [left] at (0.22,-0.5);
        \end{tikzpicture}\\[-0.1cm]
     \Xhline{2\arrayrulewidth}
\end{tabular}
\caption{Examples of cloud masks for which the majority of responders (79.2\%, 63.4\%, and 89.1\%) said that a)\,none of the masks is of enough quality (\Jaccard: 0.894, \Jaccard$^{\rm PP}$: 0.941), b)\,mask obtained without post processing (PP) is better (\Jaccard: 0.594, \Jaccard$^{\rm PP}$: 0.614), and c)~the mask with PP is better (\Jaccard: 0.671, \Jaccard$^{\rm PP}$: 0.693). We also show $\Delta={\rm \Jaccard}^{\rm PP}-{\rm \Jaccard}$.}\label{fig:mos}
\end{figure}
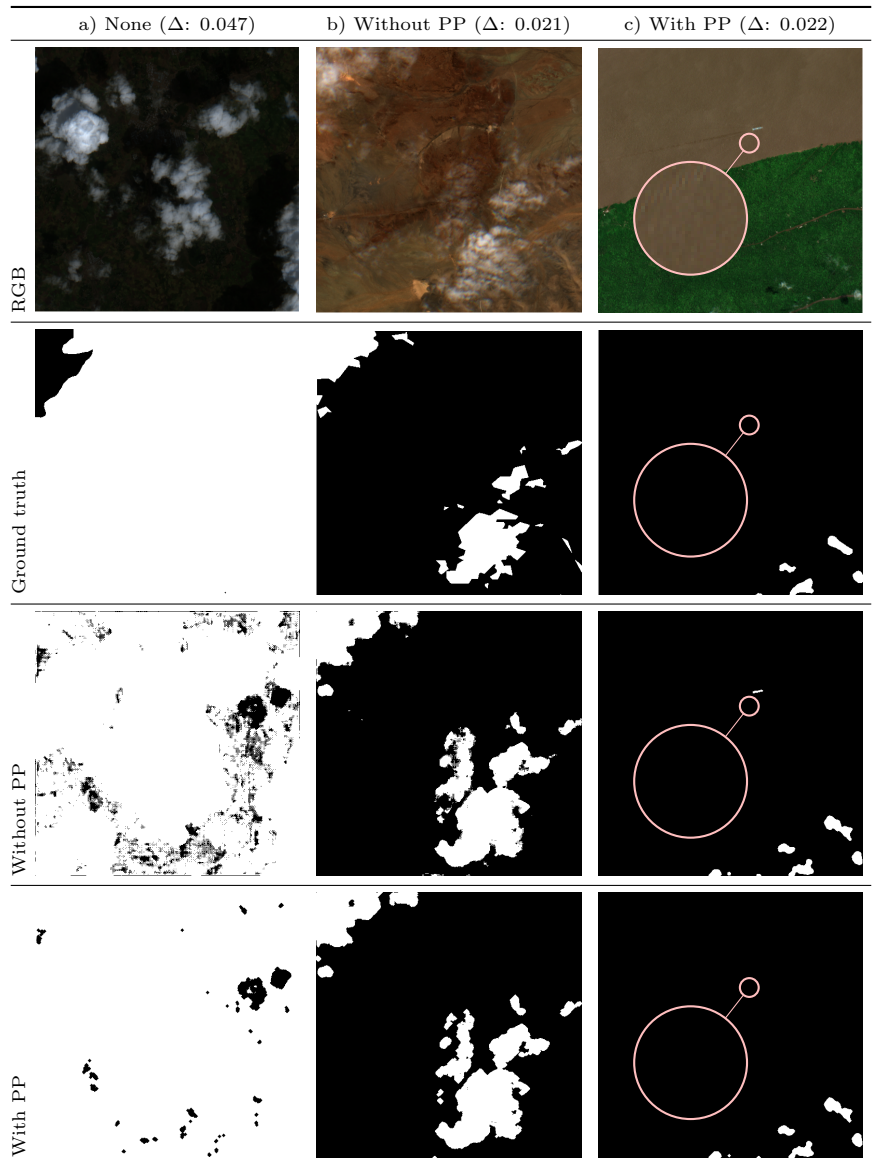

The MOS results (Table~\ref{tab:mos}) indicate a strong disagreement between the responders in all scenarios, i.e.,~across the images for which the mask A or B (without and with post processing) had a larger \Jaccard, and across all (15) images. Although indeed the majority of participants annotated mask A as better than B with a significant margin in the first case (larger \Jaccard\, for mask A, first row), it is not that evident in the second scenario (larger \Jaccard\, for mask B, second row). In all scenarios, approximately half of the responders did not select a single mask as better (the sum of ``Both'' and ``None'' were 53.06\%, 45.38\%, and 49.99\%) which shows that spotting a difference across the cloud masks was challenging to humans, or both masks were perceived as the one of insufficient quality. Further qualitative analysis in Figure~\ref{fig:mos} highlights the questionable quality of the GT (Figure~\ref{fig:mos}a) which may notably affect the training process---although \Jaccard\, is large in this case, the cloud mask is unacceptable. In Figure~\ref{fig:mos}b, the larger \Jaccard\, was in contradiction to the visual investigation (63.4\% participants annotated mask A as better, whereas only 8.9\% selected mask B), and only Figure~\ref{fig:mos}c presents the scene in which \Jaccard\, was in line with MOS, as a tiny false-positive object was pruned in post processing. We can observe that \Jaccard\, can be misleading, as it is affected by the GT quality. MOS provided the evidence that small differences in \Jaccard\,are often not perceivable (``Both'' in Table~\ref{tab:mos}). Such minor improvements may not be worthwhile in real-world missions due to the development cost/segmentation quality trade-offs. 

\begin{figure}[ht!]
\centering
\hspace*{-0.7cm}
\includegraphics[width=1.1\columnwidth]{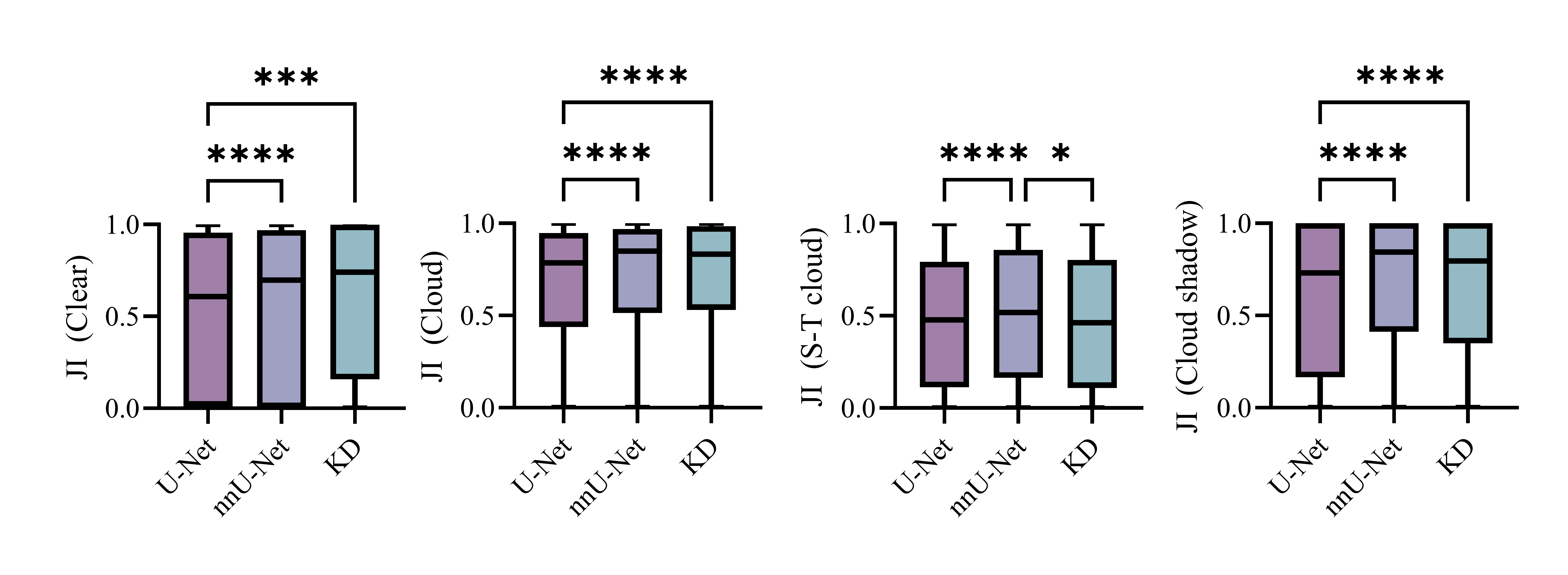}\\[-0.3cm]
\caption{The differences in \Jaccard\,across classic U-Nets, nnU-Nets and the U-Nets trained with KD (****: $p<0.0001$, ***: $p<0.001$, **: $p<0.01$, *: $p<0.05$, according to Friedman's test with post-hoc Dunn's) show that KD significantly improves vanilla U-Nets. }\label{fig:boxplots}
\end{figure}

In Figure~\ref{fig:boxplots} and Table~\ref{tab:kd}, we present \Jaccard~obtained for 872 test images which emulate the $\Phi$-Sat-2 on-board image acquisition conditions in KZ obtained using a vanilla U-Net optimized for on-board processing\footnote{We removed the second convolutional layer together with the corresponding batch normalization, as it was extremely time-consuming during the inference---this operation increased FPS from 1.63 to 5.57 of the U-Net.} (450k parameters~\cite{9554170}) trained from scratch without KD, nnU-Nets (an ensemble of five models, approx. 125M parameters), and the same U-Net architecture trained with KD for distilling the knowledge from nnU-Nets. The results show that KD significantly enhances vanilla U-Nets and leads to U-Nets working on par with much larger nnU-Nets in terms of their segmentation ability, with a dramatic reduction of the size of the ensemble of nnU-Nets (approx. $280\times$ reductions is achieved, from 150M down to 450k parameters). Also, we benchmarked the U-Net for on-board processing on Intel Movidius Myriad-2, connected by the USB port to a PC computer---for a 19.4 $\times$ 19.4 km scene, the inference took 12.2\,s without any drop in segmentation performance when compared to its GPU counterpart ($p<0.01$). This U-Net architecture will fly on-board $\Phi$-Sat-2 and will be a smart image selection step for other applications, which further proves its on-board capabilities.

\begin{table}[ht!]
\centering
\scriptsize
  \caption{The results (\Jaccard) aggregated for the KZ test set (we report the mean $\mu$, median $\mathcal{M}$ and the 95\% Confidence Interval). The best results are boldfaced, whereas the second best are underlined.}
  \label{tab:kd}
  \renewcommand{\tabcolsep}{1.3mm}
  \begin{tabular}{cccccc}
    \Xhline{2\arrayrulewidth}
    Model & Metric & Clear & Cloud & S-T cloud & Cloud shadow\\
    \hline
    & $\mu$ & 0.533	& 0.658 & \uline{0.470} & 0.607\\
    U-Net & 95\% CI & 0.506--0.560 & 0.635--0.681 & 0.446--0.493 & 0.581--0.634\\
    & $\mathcal{M}$ & 0.607 & 0.786 & \uline{0.477} & 0.732\\
    \hline
    & $\mu$ & \uline{0.548}	& \uline{0.694} & \textbf{0.510} & \textbf{0.680}\\
    nnU-Net & 95\% CI & 0.520--0.576 & 0.671--0.717 & 0.486--0.534 & 0.654--0.706\\
    & $\mathcal{M}$ & \uline{0.698} & \textbf{0.850} & \textbf{0.518} & \textbf{0.845}\\
    \hline
    \multirowcell{3}{U-Net\\with KD}& $\mu$ & \textbf{0.593}	& \textbf{0.698} & \uline{0.470} & \uline{0.659}\\
     & 95\% CI & 0.567--0.620 & 0.676--0.720 & 0.446--0.493 & 0.633--0.684\\
     & $\mathcal{M}$ & \textbf{0.740} & \uline{0.832} & 0.463 & \uline{0.796}\\
    \Xhline{2\arrayrulewidth}
\end{tabular}
\end{table}

\section{Conclusions}

We tackled the cloud detection task in a fully data-driven manner, and proposed to utilize the nnU-Nets in this context. We showed that nnU-Nets offer the state-of-the-art performance with zero user intervention---to the best of our knowledge, they have never been used in satellite image analysis before. Our experiments, performed over large-scale sets showed that the suggested pipeline can be easily used for new image data, and it delivers the performance which is on par with or outperforming the hand-crafted algorithms. Our nnU-Net was ranked within the top 7\% approaches in the On Cloud N: Cloud Cover Detection Challenge which attracted almost 850 teams. To make use of large-capacity nnU-Nets in on-board processing, we utilized knowledge distillation to train ca.\,280$\times$ smaller U-Nets (with only 450k parameters), while still benefiting from nnU-Nets---this approach significantly improved vanilla U-Nets in multi-class cloud segmentation. Such a compact U-Net model will fly on board the $\Phi$-Sat-2 mission.

\section*{Acknowledgements}

This work was funded by ESA, partly via a feasibility study for CHIME mission, and via Intuition-1 focused GENESIS project supported by $\Phi$-lab (https://philab.esa.int/). MK and JN were supported by the Silesian University of Technology grant for maintaining and developing research potential.

\bibliography{ref_all}

\begin{thebibliography}{15}
\expandafter\ifx\csname natexlab\endcsname\relax\def\natexlab#1{#1}\fi
\providecommand{\url}[1]{\texttt{#1}}
\providecommand{\href}[2]{#2}
\providecommand{\path}[1]{#1}
\providecommand{\DOIprefix}{doi:}
\providecommand{\ArXivprefix}{arXiv:}
\providecommand{\URLprefix}{URL: }
\providecommand{\Pubmedprefix}{pmid:}
\providecommand{\doi}[1]{\href{http://dx.doi.org/#1}{\path{#1}}}
\providecommand{\Pubmed}[1]{\href{pmid:#1}{\path{#1}}}
\providecommand{\bibinfo}[2]{#2}
\ifx\xfnm\relax \def\xfnm[#1]{\unskip,\space#1}\fi
\bibitem[{Jeppesen et~al.(2019)}]{JEPPESEN2019247}
\bibinfo{author}{J.~Jeppesen}, et~al.,
\newblock \bibinfo{title}{A cloud detection algorithm for satellite imagery
  based on deep learning},
\newblock \bibinfo{journal}{Remote Sens. Environ.} \bibinfo{volume}{229}
  (\bibinfo{year}{2019}) \bibinfo{pages}{247 -- 259}.
\bibitem[{Li et~al.(2021)}]{Li2021}
\bibinfo{author}{L.~Li}, et~al.,
\newblock \bibinfo{title}{A review on deep learning techniques for cloud
  detection methodologies and challenges},
\newblock \bibinfo{journal}{Signal Image Video Proc.} \bibinfo{volume}{15}
  (\bibinfo{year}{2021}) \bibinfo{pages}{1527--1535}.
\bibitem[{{Mohajerani} and {Saeedi}(2019)}]{38-cloud-1}
\bibinfo{author}{S.~{Mohajerani}}, \bibinfo{author}{P.~{Saeedi}},
\newblock \bibinfo{title}{{Cloud-Net:} an end-to-end cloud detection algorithm
  for {Landsat 8} imagery},
\newblock in: \bibinfo{booktitle}{Proc. IGARSS}, \bibinfo{year}{2019}, pp.
  \bibinfo{pages}{1029--1032}.
\bibitem[{Mahajan and Fataniya(2020)}]{Mahajan2020}
\bibinfo{author}{S.~Mahajan}, \bibinfo{author}{B.~Fataniya},
\newblock \bibinfo{title}{Cloud detection methodologies: variants and
  development---a review},
\newblock \bibinfo{journal}{Complex Intell. Syst.} \bibinfo{volume}{6}
  (\bibinfo{year}{2020}) \bibinfo{pages}{251--261}.
\bibitem[{Nalepa et~al.(2021)}]{rs13081532}
\bibinfo{author}{J.~Nalepa}, et~al.,
\newblock \bibinfo{title}{Towards on-board hyperspectral satellite image
  segmentation: Understanding robustness of deep learning through simulating
  acquisition conditions},
\newblock \bibinfo{journal}{Remote Sens.} \bibinfo{volume}{13}
  (\bibinfo{year}{2021}).
\bibitem[{Domnich et~al.(2021)}]{rs13204100}
\bibinfo{author}{M.~Domnich}, et~al.,
\newblock \bibinfo{title}{{KappaMask: AI-Based Cloudmask Processor for
  Sentinel-2}},
\newblock \bibinfo{journal}{Remote Sens.} \bibinfo{volume}{13}
  (\bibinfo{year}{2021}).
\bibitem[{Mohajerani et~al.(2018)}]{38-cloud-2}
\bibinfo{author}{S.~Mohajerani}, et~al.,
\newblock \bibinfo{title}{A cloud detection algorithm for remote sensing images
  using fully convolutional neural networks},
\newblock in: \bibinfo{booktitle}{Proc. MMSP}, \bibinfo{year}{2018}, pp.
  \bibinfo{pages}{1--5}.
\bibitem[{Zhu et~al.(2015)}]{ZHU2015269}
\bibinfo{author}{Z.~Zhu}, et~al.,
\newblock \bibinfo{title}{{Improvement and expansion of the Fmask algorithm:
  cloud, cloud shadow, and snow detection for Landsats 4–7, 8, and Sentinel 2
  images}},
\newblock \bibinfo{journal}{Remote Sens. Environ.} \bibinfo{volume}{159}
  (\bibinfo{year}{2015}) \bibinfo{pages}{269--277}.
\bibitem[{Yanan et~al.(2020)}]{Yanan_2020}
\bibinfo{author}{G.~Yanan}, et~al.,
\newblock \bibinfo{title}{Cloud detection for satellite imagery using deep
  learning},
\newblock \bibinfo{journal}{J. Phys. Conf. Ser.} \bibinfo{volume}{1617}
  (\bibinfo{year}{2020}) \bibinfo{pages}{012089}.
\bibitem[{Salinas et~al.(2021)}]{10.1007/978-3-030-86517-7_28}
\bibinfo{author}{P.~Salinas}, et~al.,
\newblock \bibinfo{title}{{Automated Machine Learning for Satellite Data:
  Integrating Remote Sensing Pre-trained Models into AutoML Systems}},
\newblock in: \bibinfo{booktitle}{Proc. ECML PKDD},
  \bibinfo{publisher}{Springer}, \bibinfo{year}{2021}, pp.
  \bibinfo{pages}{447--462}.
\bibitem[{Ziaja et~al.(2021)}]{rs13193981}
\bibinfo{author}{M.~Ziaja}, et~al.,
\newblock \bibinfo{title}{Benchmarking deep learning for on-board space
  applications},
\newblock \bibinfo{journal}{Remote Sensing} \bibinfo{volume}{13}
  (\bibinfo{year}{2021}).
\bibitem[{Gou et~al.(2021)Gou, Yu, Maybank, and Tao}]{Gou2021}
\bibinfo{author}{J.~Gou}, \bibinfo{author}{B.~Yu}, \bibinfo{author}{S.~J.
  Maybank}, \bibinfo{author}{D.~Tao},
\newblock \bibinfo{title}{Knowledge distillation: A survey},
\newblock \bibinfo{journal}{International Journal of Computer Vision}
  \bibinfo{volume}{129} (\bibinfo{year}{2021}) \bibinfo{pages}{1789--1819}.
\bibitem[{Isensee et~al.(2021)}]{isensee_nnu-net_2021}
\bibinfo{author}{F.~Isensee}, et~al.,
\newblock \bibinfo{title}{{nnU}-{Net}: a self-configuring method for deep
  learning-based biomedical image segmentation},
\newblock \bibinfo{journal}{Nature Methods} \bibinfo{volume}{18}
  (\bibinfo{year}{2021}) \bibinfo{pages}{203--211}.
\bibitem[{Hinton et~al.(2015)}]{https://doi.org/10.48550/arxiv.1503.02531}
\bibinfo{author}{G.~Hinton}, et~al.,
\newblock \bibinfo{title}{Distilling the knowledge in a neural network}
  (\bibinfo{year}{2015}).
\bibitem[{Grabowski et~al.(2021)}]{9554170}
\bibinfo{author}{B.~Grabowski}, et~al.,
\newblock \bibinfo{title}{Towards robust cloud detection in satellite images
  using {U-Nets}},
\newblock in: \bibinfo{booktitle}{Proc. IGARSS}, \bibinfo{year}{2021}, pp.
  \bibinfo{pages}{4099--4102}.

\end{thebibliography}

\end{document}